\pdfoutput=1

\documentclass[11pt]{article}

\usepackage[]{acl}

\usepackage{times}
\usepackage{latexsym}

\usepackage[T1]{fontenc}

\usepackage[utf8]{inputenc}

\usepackage{microtype}

\usepackage{amssymb}
\usepackage{booktabs}
\usepackage{xcolor} 
\definecolor{LightGray}{gray}{0.9}
\definecolor{x11gray}{rgb}{0.75, 0.75, 0.75}
\definecolor{lavendergray}{rgb}{0.77, 0.76, 0.82}
\definecolor{lightgray}{rgb}{0.83, 0.83, 0.83}
\definecolor{rred}{HTML}{C0504D}
\definecolor{ggreen}{HTML}{9BBB59}
\usepackage{graphicx}
\usepackage{makecell}
\usepackage{xcolor}
\usepackage{pgfplots}
\usepackage{tikz}

\usepackage{url}
\usepackage{xurl}

\title{EHRKit: A Python Natural Language Processing Toolkit for \\Electronic Health Record Texts}

\author{Irene Li, Keen You, Yujie Qiao, Lucas Huang,  \\ \textbf{Chia-Chun Hsieh, Benjamin Rosand, Jeremy Goldwasser, Dragomir Radev}
\\
Yale University
}

\begin{document}
\maketitle
\begin{abstract}
The Electronic Health Record (EHR) is an essential part of the modern medical system and impacts healthcare delivery, operations, and research. Unstructured text is attracting much attention despite structured information in the EHRs and has become an exciting research field. The success of the recent neural Natural Language Processing (NLP) method has led to a new direction for processing unstructured clinical notes. In this work, we create a python library for clinical texts, EHRKit. This library contains two main parts: MIMIC-III-specific functions and task-specific functions. The first part introduces a list of interfaces for accessing MIMIC-III NOTEEVENTS data, including basic search, information retrieval, and information extraction. The second part integrates many third-party libraries for up to 12 off-shelf NLP tasks such as named entity recognition, summarization, machine translation, etc.

\end{abstract}

\section{Introduction}

With the rising trend of Electronic Health Records (EHRs), massive unstructured texts (i.e., clinical and admission notes) are being created in the healthcare system. It is very important to process such data for secondary usage \cite{xiao2018oppor}. The main obstacle is the processing and understanding of the unstructured text. Natural Language Processing (NLP) techniques have been applied to deal with such texts \cite{li2021neuraln,shickel2018deep,aiad2018survey}. Especially, deep learning-based methods achieved great success in some existing NLP tasks in the biomedical and clinical literature, such as text classification \cite{zhou2021automatic,li2020icd, li2019neural,hughes2017medical}, named entity recognition \cite{song2021deep}, text segmentation \cite{badjatiya2018attention}, medical language translation and generation \cite{weng2019unsupervised,abacha2019summarization} and many others.

Following the success of BERT \cite{devlin-etal-2019-bert}, researchers developed BERT-based models trained on the clinical literature, such as BioBERT \cite{lee2020biobert} and ClinicalBERT\cite{clinicalbert}. 
With such advanced neural models, there is a need for a user-friendly programming interface that can support a variety of downstream tasks. Some existing libraries and toolkits are designed for bioinformatical and clinical needs, including the biomedical and clinical model packages of Stanza \cite{zhang2021biomedical}, SciFive \cite{phan2021scifive}, UmlsBERT \cite{umlsbert}, MIMIC-Extract \cite{wang2020mimic} and so on. However, we noticed a need to integrate more existing libraries with much broader coverage of clinical and biomedical NLP tasks in our toolkit. Besides, based on the analysis from \citet{li2021neuraln}, there is limited research on generation tasks for EHR unstructured text, i.e., machine translation. Thus, we provide a pretrained model for clinical text machine translation in our toolkit, which supports three languages.

\begin{figure*}[t]
    \centering
    \includegraphics[width=0.9\textwidth]{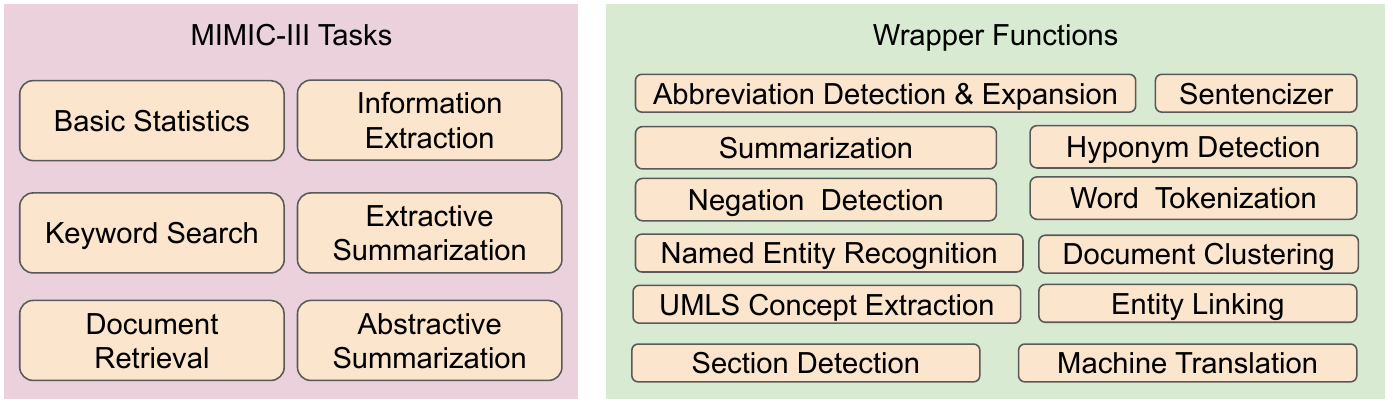}
    \caption{EHRKit Architecture.}
    \label{fig:ehrkit}
\end{figure*}

Our contributions are: 1) First, we propose EHRKit, a Python NLP toolkit for EHR unstructured texts. This toolkit contains two main components: general API functions and MIMIC-specific functions. It is user-friendly, with easy installation and quick start tutorials. 2) Second, to fill the gap in text generation for clinical texts, we release pretrained machine translation models in three languages. Besides, we evaluate existing pretrained models for summarization in a biomedical and clinical scenario. We release this toolkit and pretrained models publicly available at \url{placeholder.link}.

\section{System Design and Architecture}

We show our EHRKit architecture in Fig.~\ref{fig:ehrkit}. It consists of two modules, namely MIMIC-III Tasks and Wrapper Functions. 

\subsection{MIMIC-III Tasks} 

We include some basic NLP functions for MIMIC-III NOTEEVENTS text data \cite{johnson2016mimic}. 

\begin{itemize}
\itemsep0em 
    \item Basic statistical functions support counting for number of patients and number of documents, number of sentences and so on. \item Information Extraction provides helpful interfaces for investigating the data, i.e.,  phrases extraction, and abbreviation term extraction from a given record ID. We applied Phrase-At-Scale\footnote{\url{https://github.com/kavgan/phrase-at-scale}} for this function. 
    \item Keyword Search allows users to search a record by a keyword.
    \item Document Retrieval allows users to search a record by ID.
    \item Extractive and Abstractive Summarization: Na\"ive Bayes \cite{ramanujam2016automatic} and DistilBART \cite{shleifer2020pre} for text summarization.
\end{itemize}

\subsection{Wrapper Functions}
This module integrates many third-party libraries and supports up to 12 functionalities for any free-text inputs. 

\begin{itemize}
\itemsep0em 
    \item Abbreviation Detection and Expansion: finds abbreviation and its expansions. Function imported from ScispaCy\footnote{\url{https://allenai.github.io/scispacy/}}.
    
    \item Sentencizer: detects sentence boundaries. We support four approaches: PyRuSH\footnote{\url{https://github.com/jianlins/PyRuSH}.}, Stanza, ScispaCy and Stanza Biomed.
    
    \item Hyponym Detection: finds the hyponyms of the recognized entities in the input text. Function imported from scispaCy.
    
    \item Negation Detection: detects negation in a sentence, imported from medspaCy \cite{eyre2021medspacy}
    
    \item Word Tokenization: tokenizes a sentence into a list of words. Function imported from medspaCy.
    
    \item Named Entity Recognition: finds named entities, part-of-speech and universal morphological features, and dependencies of an input record. Function imported from Stanza \cite{zhang2021biomedical}.
    
    \item Document Clustering: given the query record, selects $k$ documents from supporting records that are most similar to the main record (K-Means clustering), measured by embedded document using pretrained BERT model \cite{devlin-etal-2019-bert}.
    
    \item UMLS Concept Extraction: matches the UMLS concept for the input text. Function imported from medspaCy \cite{eyre2021medspacy}.
    
    \item Entity Linking: finds named entities, negation entities, and linked entities in the input text. Function imported from scispaCy.
    
    \item Section Detection: rule-based method for detecting section (i.e., \textit{allergies}, \textit{history}). Function imported from medspaCy \cite{eyre2021medspacy}.
    
    \item Machine Translation: translates English texts into 17 target languages. We applied the existing MarianMT model\footnote{\url{https://huggingface.co/docs/transformers/model_doc/marian}}, as well as our own fine-tuned models. 
    
    \item Summarization: we support both extractive and abstractive summarization methods. We integrated TextRank \cite{mihalcea2004textrank}, pretrained BART \cite{lewis2019bart}, T5 \cite{raffel2020explore} and SciFive summarization libraries. We also allow single and multiple documents as the input. 
\end{itemize}



\subsection{Other similar libraries}

\textbf{MIMIC-Extract}\footnote{\url{https://github.com/MLforHealth/MIMIC_Extract}}A pipeline for preprocessing and presenting data from MIMIC-III dataset.  It provides features for data analysis, including extraction of clinical events like mortality from free text.

\textbf{ScispaCy} \cite{neumann-etal-2019-scispacy} A tool that adapts SpaCy’s models to process scientific, biomedical, and clinical text. It supports multiple methods for tokenization, part of speech tagging, dependency parsing, and named entity recognition. 

\textbf{medspaCy} Based on the spaCy framework,  medspaCy \cite{eyre2021launching} is a clinical NLP python library that provides both rule-based and machine learning-based methods for processing clinical text. It supports methods for various clinical applications such as UMLS Mapping (rule-based), Section Detection, Sentence Detection, Contextual Analysis and Visualization on entities.


\begin{table}[t]
\centering
\begin{tabular}{lccccc} \toprule
               & {\footnotesize MIMIC} & {\footnotesize Neu}  & {\footnotesize MT} & {\footnotesize Summ}  \\
              \midrule
{\footnotesize MIMIC-Extract}     & {\footnotesize \checkmark}  &          &      &         &             \\
{\footnotesize ScispaCy}        &       &   {\footnotesize \checkmark}           &     &              \\
{\footnotesize medspaCy}        &       &   {\footnotesize \checkmark}          &     &              \\
{\footnotesize Stanza Biomed}    &     &      {\footnotesize \checkmark}     &    &          \\
{\footnotesize SciFive}        &     &  {\footnotesize \checkmark}       &      &    {\footnotesize \checkmark}      \\
{\footnotesize EHRKit (ours)} &  {\footnotesize \checkmark}      &  {\footnotesize \checkmark}   &   {\footnotesize \checkmark}  & {\footnotesize \checkmark}     \\
\bottomrule
\end{tabular}
\caption{A comparison with other similar python toolkits. \texttt{MIMIC}: MIMIC Related. \texttt{Neu}: Neural Methods. \texttt{MT}: Machine Translation. \texttt{Summ}: Summarization. }
\label{tab:comparison}
\end{table}

\textbf{Stanza Biomed} \cite{zhang2021biomedical} A set of tools for statistical, neural, and rule-based problems in computational linguistics. Its software provides a simple interface for NLP tasks. It is a widely used Python library for processing clinical texts. It provides nearly state-of-the-art performance using neural networks on tasks including tokenization, sentence segmentation, part of speech (POS) tagging, lemmatization, and dependency parsing.

\textbf{SciFive} \cite{phan2021scifive} A pretrained neural language model for biomedical domain. Fine-tuned on PubMed Abstract \footnote{\url{https://pubmed.ncbi.nlm.nih.gov/}} and PubMed Central (PMC) \footnote{\url{https://www.ncbi.nlm.nih.gov/pmc}}, it outperformed similar models including BioBERT and T5 on multiple NLP tasks: named entity relation, relation extraction, natural language inference, and question answering.



Tab.~\ref{tab:comparison} lists our EHRKit and other similar toolkits. We compare from different perspectives by focusing on the functionalities. \textbf{MIMIC Related}: if supports MIMIC-related functions. We consider MIMIC an essential data source that plays an important role in research. We can find that only MIMIC-Extract and EHRKit support these related functions, and users can apply them directly to the MIMIC data. \textbf{Neural Methods}: if this toolkit supports neural methods and embedding methods. As we can see, the majority contain such features. \textbf{Machine Translation} and \textbf{Summarization}: if this toolkit supports (neural) generation tasks like machine translation and summarization. In this case, only SciFive and EHRKit support such features. Our toolkit provides diverse functionalities and is easy to use based on these perspectives.

\begin{table}[t]
\centering
\begin{tabular}{crrr}
\toprule
        Lang. Pair            & Total     & Train     & Test    \\ \midrule
\makecell{en $\rightarrow$ es}  & 790,915   & 672,276   & 111,779 \\ 
\makecell{en$\rightarrow$fr}   & 2,812,305 & 2,390,458 & 407,388 \\ 
\makecell{en$\rightarrow$ ro} & 1,165,092 & 990,327   & 161,936 \\ \bottomrule
\end{tabular}
\caption{Data statistics for machine translation: we apply UFAL and select the overlapped target language pairs for our experiments.   }
\label{tab:stats_ufal}
\end{table}

\begin{table*}[th]
\centering
\small
\begin{tabular}{cccccccc}
\toprule
 & \multicolumn{3}{c}{PubMed}  &  \multicolumn{3}{c}{MIMIC-CXR} \\
 & R-1     & R-2     & R-L  &  R-1     & R-2     & R-L   \\ \midrule
Pegasus \cite{zhang2020pegasus}  & 
45.97&20.15&28.25 
& 22.49 &	11.57&	20.35
\\ 
BigBird \cite{zaheer2020big}   & 46.32&	20.65	& \textbf{42.33} &
38.99 &	29.52 &	38.59
\\ 
BART \cite{lewis2019bart}  & \textbf{48.35} & \textbf{21.43} & 36.90 &
\textbf{41.70} &	\textbf{32.93} &	\textbf{41.16}
\\
SciFive \cite{phan2021scifive}  & - & - & - &
35.41 &	26.48 & 	35.07
 \\ 
 \bottomrule
\end{tabular}
\caption{Single document summarization evaluation: we evaluate selected models and report ROUGE-1, ROUGE-2 and ROUGE-L. We adapted a few results reported by \citet{rohde2021hierarchical}.  SciFive applied PubMed during pretraining, so we did not provide the evaluation.   }
\label{tab:sum}
\end{table*}











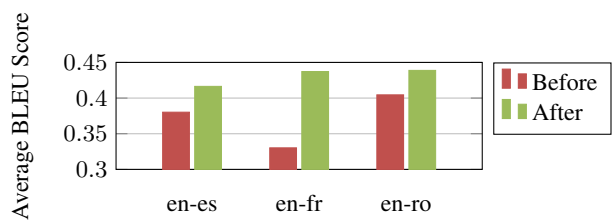
\begin{figure}[t]
\small
\begin{tikzpicture}
    \begin{axis}[
        width  = 0.4*\textwidth,
        height = 3cm,
        major x tick style = transparent,
        ybar=0pt,
        bar width=10pt,
        ymajorgrids = true,
        ylabel = {Average BLEU Score},
        symbolic x coords={en-es,en-fr,en-ro},
        xticklabel style={text height=2ex},
        xtick = data,
        scaled y ticks = false,
        ymin=0.3,
        ymax=0.45,
        ybar, enlarge x limits ={abs=1cm},
        legend style={at={(0.5,0.97)}, anchor=north, legend cell align=left,     align=left, draw=black},
         legend pos=outer north east,
         legend cell align=left,
    ]

        \addplot[style={rred,fill=rred,mark=none}]
            coordinates {(en-es,0.3802) (en-fr,0.3302) (en-ro,0.4045)};

        \addplot[style={ggreen,fill=ggreen,mark=none}]
            coordinates {(en-es,0.4164) (en-fr,0.4372) (en-ro,0.4388)};

        \legend{Before, After}
    \end{axis}
\end{tikzpicture}

    \caption{BLEU score: before and after pretraining.}
    \label{tab:bleu}
    \vspace{-3mm}
\end{figure}

\begin{table}[t]
\centering
\begin{tabular}{clll} 

\toprule
Dataset & Train & Valid & Test \\
\midrule
PubMed & 112k & 6.6k & 6.7k \\
MIMIC-CXR & 89,544 & 2000 & 2000 \\
\bottomrule
\end{tabular}
\caption{Data statistics for PubMed and MIMIC-CXR summarization datasets. Words are counted before tokenization.}
\label{table:datastats}
\vspace{-5mm}
\end{table}

\section{Performance Evaluation}

\subsection{Machine Translation} 

We report the performance of the Machine Translation function from our EHRKit and compare it with the baseline model, MarianMT. Our training sets and test sets are obtained from the UFAL Medical Corpus\footnote{\url{https://ufal.mff.cuni.cz/ufal_medical_corpus}}. These data are from various medical text sources, such as titles of medical Wikipedia articles, medical term-pairs, patents, and documents from the European Medicines Agency \cite{Braune2018D1}. We evaluate language pairs, including English (en) to Spanish (es), English to French (fr), and English to Romanian (ro), as those are the three language pairs that EHRKit and UFAL mutually support. 

For data pre-processing, we first exclude general domain data from UFAL, such as parliament proceedings. Next, we randomly shuffle the medical-domain corpora and split it into two parts by 85\% and 15\%, as our training set and test set, respectively. For each language pair, we use all of the available parallel data. Tab. \ref{tab:stats_ufal} summarizes the number of sentences that we extract from UFAL. 




Subsequently, we evaluate model performance using the BLEU score \cite{papineni-etal-2002-bleu}. We finetune our model with the training sets (After) and compare it with the baseline model,  MarianMT (Before). As shown in Fig.~\ref{tab:bleu}, our model improves significantly after finetuning, with an average 16.80\% increase in BLEU score. Among the three selected language pairs, we can observe that English to French has the best improvement - it achieves 32.40\% performance gain. We conjecture that this occurs because we have significantly more training data in English to French.





\subsection{Summarization}


We explore the performance of some pre-existing summarization models from the broader NLP field when applied to biomedical literature and clinical data. Given the difficulty in obtaining summarization corpora from clinical settings, we opted to use the existing PubMed \cite{cohan-etal-2018-discourse} and MIMIC-CXR \cite{johnson2019mimic} as our primary datasets.

The PubMed dataset comprises 133k biomedical scientific publications sourced from the PubMed database. In this dataset, each input document is a scientific article, with the corresponding abstract serving as the reference summary. MIMIC-CXR is a de-identified, Protected Health Information removed dataset of chest radiographs, 
We use a subset from the MIMIC-CXR for the MEDIQA 2021 Radiology report summarization shared task \cite{delbrouck-etal-2021-qiai}. Since we did not obtain the ground truth of the original test data, we applied the validation as our test set, and left out 2000 examples from the original training set for validation. 
We show the statistics in Tab.~\ref{table:datastats}.

In Table~\ref{tab:sum}, we assess a selection of pre-trained abstractive methods using the ROUGE metric \cite{lin2004rouge}. Among the four models evaluated, BART \cite{lewis2019bart}, though trained with a general purpose, demonstrates a high R-1 score. However, BigBird \cite{zaheer2020big} also showed competitive performance on R-L in PubMed. These results highlight the complexity of identifying a single superior model for our specific use case, even though SciFive was specifically pre-trained for this task. Future research could focus on enhancing the automatic summarization of biomedical and clinical texts.





\section{Conclusion}
In this work, we propose a python library for clinical texts, EHRKit. This toolkit contains two main components: general API functions and MIMIC-specific functions. 
In the future, we will investigate more EHR-NLP tasks including machine translation for more languages, multi-document summarization and question answering \cite{li2021neuraln}. Besides, we plan to investigate better-performed NLP models for these tasks, for example, BERT-based models \cite{lee2020biobert,li2021improving}.

\bibliography{anthology,custom}
\bibliographystyle{acl_natbib}
\newpage
\appendix

\section{Limitations}
As our work is a tool for processing clinical texts, we do not propose model-based novelty as one of our main contributions. Users may find that we conducted evaluations and built high-level user interfaces instead of proposing new models.

As EHRKit relies on many other existing libraries, we suggest that users install compatible and correct versions for robust usage. 

\section{Potential Risk}
This work is an open-source tool for clinical text processing. We did not use any user-sensitive data for training or testing, and this tool does not contain any related functionalities. Users should avoid using such data as inputs.

\section{Experiments}

Our models were trained on a 4 Nvidia 3090 GPUs with a batch size of 8. We train all of our models using Adagrad with 0.15 learning rate and have an accumulator of 0.1.

\subsection{Machine Translation}
The training time varies on language pairs. The total trial, training and evaluation time is about 60 ours. 

\subsection{Summarization}
During training we are regularly measuring the loss and the ROUGE-1 F-score on the validation set of the dataset in order to monitor the learning of our model. We end the training when the validation loss stops improving. The overall trial, training and evaluation time is about 20 hours. 








\end{document}